\documentclass{article}
\usepackage{ijcai22}

\usepackage{times}
\usepackage{soul}
\usepackage{url}
\usepackage[hidelinks]{hyperref}
\usepackage[utf8]{inputenc}
\usepackage[small]{caption}
\usepackage{graphicx}
\usepackage{amsmath}
\usepackage{amsthm}
\usepackage{booktabs}
\usepackage{algorithm}
\usepackage{algorithmic}
\usepackage{amssymb}
\usepackage{multicol}
\usepackage{multirow}
\usepackage{booktabs}
\usepackage{threeparttable}
\usepackage{array}
\usepackage{booktabs}
\usepackage{bbding}
\usepackage{diagbox}
\usepackage{hyperref}

\title{MMNet: Muscle Motion-guided Network for Micro-expression Recognition}

\author{
Hanting Li\and Mingzhe Sui\and Zhaoqing Zhu\And Feng Zhao\thanks{Corresponding author.}
\affiliations
University of Science and Technology of China
\emails
\{ab828658, sa20, zhaoqingzhu\}@mail.ustc.edu.cn,
fzhao956@ustc.edu.cn
}

\begin{document}

\maketitle

\begin{abstract}
  Facial micro-expressions (MEs) are involuntary facial motions revealing people’s real feelings and play an important role in the early intervention of mental illness, the national security, and many human-computer interaction systems. However, existing micro-expression datasets are limited and usually pose some challenges for training good classifiers. To model the subtle facial muscle motions, we propose a robust micro-expression recognition (MER) framework, namely muscle motion-guided network (MMNet). Specifically,  a continuous attention (CA) block is introduced to focus on modeling local subtle muscle motion patterns with little identity information, which is different from most previous methods that directly extract features from complete video frames with much identity information. Besides, we design a position calibration (PC) module based on the vision transformer. By adding the position embeddings of the face generated by the PC module at the end of the two branches, the PC module can help to add position information to facial muscle motion-pattern features for the MER. Extensive experiments on three public micro-expression datasets demonstrate that our approach outperforms state-of-the-art methods by a large margin. Code is available at \href{https://github.com/muse1998/MMNet}{https://github.com/muse1998/MMNet}.
\end{abstract}

\section{Introduction}
Facial expressions are an essential carrier for spreading human emotional information and coordinating interpersonal relationships. Most of the expressions that we see in our daily life are macro-expressions. However, spontaneous, brief, and subtle micro-expressions (MEs) can reveal people’s true feelings when people try to hide their real emotions under certain conditions \cite{ekman2009}, which makes MEs applicable to many areas such as criminal interrogation, clinical diagnosis, and human-computer interaction. Different from macro-expressions, MEs are usually accompanied by tiny facial muscle motions and last less than half a second (usually 1/25 to 1/3 second), which makes micro-expression recognition (MER) task very difficult for humans, and even more difficult for computers.

According to the features extraction methods, MER techniques can be roughly divided into two categories: handcrafted approaches and deep network-based approaches. For the former, histogram of oriented gradient (HOG), histogram of optical flow (HOOF), and local binary pattern-three orthogonal planes (LBP-TOP) are often used to extract ME features. Le Ngo et al. learnt temporal and spectral structures with sparsity constraints by processing the LBP-TOP \cite{le2016LBP-TOP}. Li et al. adopted the histogram of image gradient orientation-TOP (HIGO-TOP) for MER \cite{li2017HIGO-TOP}. Happy et al. proposed a fuzzy-based HOOF (FHOOF) feature extraction technique that only considers the muscle motion direction for MER \cite{happy2017FHOFO}. However, due to the short duration and the inconspicuous motion of MEs, handcrafted features are often unable to robustly represent the differences between different micro-expressions, which is detrimental to MER.


In recent years, with the development of deep learning technology, more and more researchers have achieved promising results by designing deep neural networks (DNNs) to handle MER tasks. For example, Gan et al. introduced a feature extractor that incorporates both the handcrafted (i.e., optical flow) and data-driven (i.e., convolutional neural networks, CNNs) features \cite{gan2019off-apex}. Song et al. proposed a three-stream convolutional neural network (TSCNN) to recognize MEs by learning ME-discriminative features in three key frames of ME videos \cite{song2019TSCNN}. Xie et al. developed a MER approach by combining action units (AUs) and emotion category labels \cite{xie2020AU-GACN}, meanwhile Lei et al. designed a graph temporal convolutional network (Graph-TCN) to extract the local muscle motion features of MEs. Xia et al. devised a framework that leverages macro-expression samples as guidance for MER \cite{Xia2020MTMNet}. Among the latest results, Xia et al. designed a macro-to-micro transformation framework by two auxiliary tasks from the spatial and temporal domains, respectively \cite{xia2021MiNetMaNet}.

Although the above methods have gradually improved the performance of automated MER algorithms in recent years. Nevertheless, most of them directly input the original video frames or the handcrafted features of the video frames into DNNs to extract features of MEs, which makes DNNs easy to lead into the identity information of the samples. Obviously, identity information that has nothing to do with expression is harmful to facial expression recognition (FER) task. This problem may have little impact on the macro-expression recognition task with abundant training data \cite{li2020survey}. However, due to the extremely cost-consuming of collecting and labeling micro-expression data, we still do not have a large-scale micro-expression dataset comparable to the macro-expression datasets (e.g., AffectNet \cite{mollahosseini2017affectnet}). MEs are mainly determined by the position of facial muscle motion and the muscle motion pattern (e.g., the slight upturn of the lips corner on both sides is likely to indicate happiness). So, the key of MER is to learn the position and pattern of facial muscle motion rather than directly learn from the entire video frames.

\begin{figure}[t]
	\centering
	\centerline{\includegraphics[width=8cm]{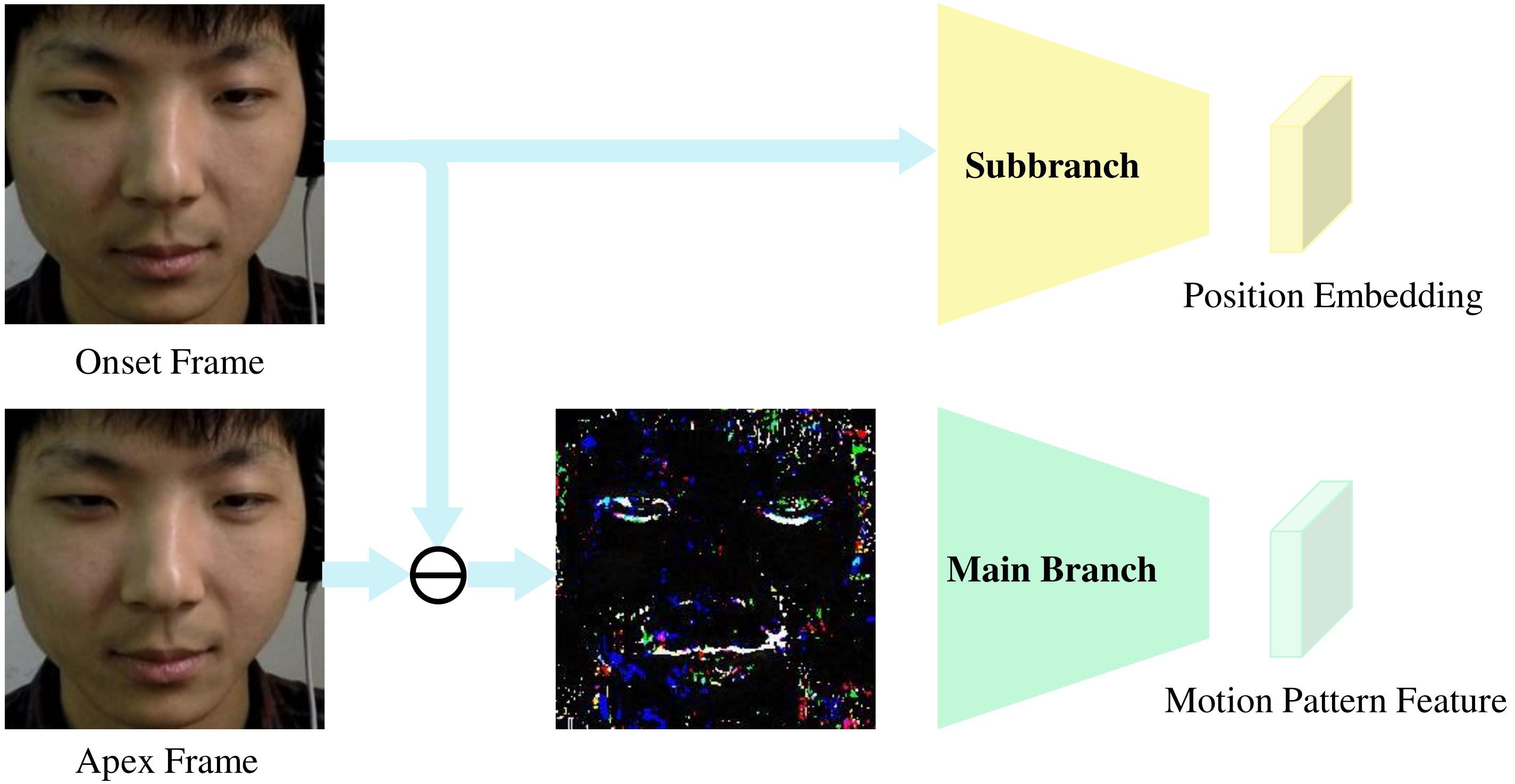}}	
	\caption{The proposed two-branch MER paradigm.}
	\label{figure1}
\end{figure}
As shown in Figure 1, we propose a new two-branch MER paradigm to deal with the two key factors mentioned above, which extracts muscle motion-pattern features from the difference between the onset frame and the apex frame through the main branch, and generates facial position embeddings from the onset frame only through the subbranch. We also give a specific realization of the proposed two-branch MER paradigm, namely muscle motion-guided network (MMNet), which mainly focuses on learning the position (e.g., lips corner and upper eyebrow) and motion patterns (e.g., raising and lowering) of facial muscle motions. Specifically, we learn the features of micro-expression video sequences by a two-branch framework. First, we introduce a new continuous attention (CA) block to learn the patterns of facial muscle motions. CA block can pay attention to the location of the motion and extract features related to motion patterns. Second, we devise a position calibration (PC) module based on vision transformer (ViT) \cite{dosovitskiy2020vit} to add robust facial position information to the learned motion-pattern features. It is worth noting that the main branch of our MMNet only models the difference between the onset frame and the apex frame, which reflects the muscle motion on the face and makes the model less affected by the identity information.

The contributions of our work are summarized as follows:
\begin{itemize}
	\item We propose a novel two-branch MER paradigm, which extracts the muscle motion-pattern features and facial position embeddings through the main branch and subbranch, respectively. Then, the two kinds of features are fused at the end of the network for classification.
	\item We devise a muscle motion-guided network (MMNet) to implement the above two-branch MER paradigm. The main branch for extracting the motion-pattern features is composed of the proposed continuous attention (CA) block, and the subbranch for generating the position embeddings is realized through the designed position calibration (PC) module.
	\item Our MMNet outperforms the state-of-the-art approaches by a large margin on three popular micro-expression datasets (i.e., CASME II, SAMM, and MMEW). Extensive experiments demonstrate the effectiveness of the proposed MMNet.
\end{itemize}

\section{Method}
In this section, we detail the two branches of our proposed MMNet for MER. 

\begin{figure*}[t]
  \centering
  \centerline{\includegraphics[width=15cm]{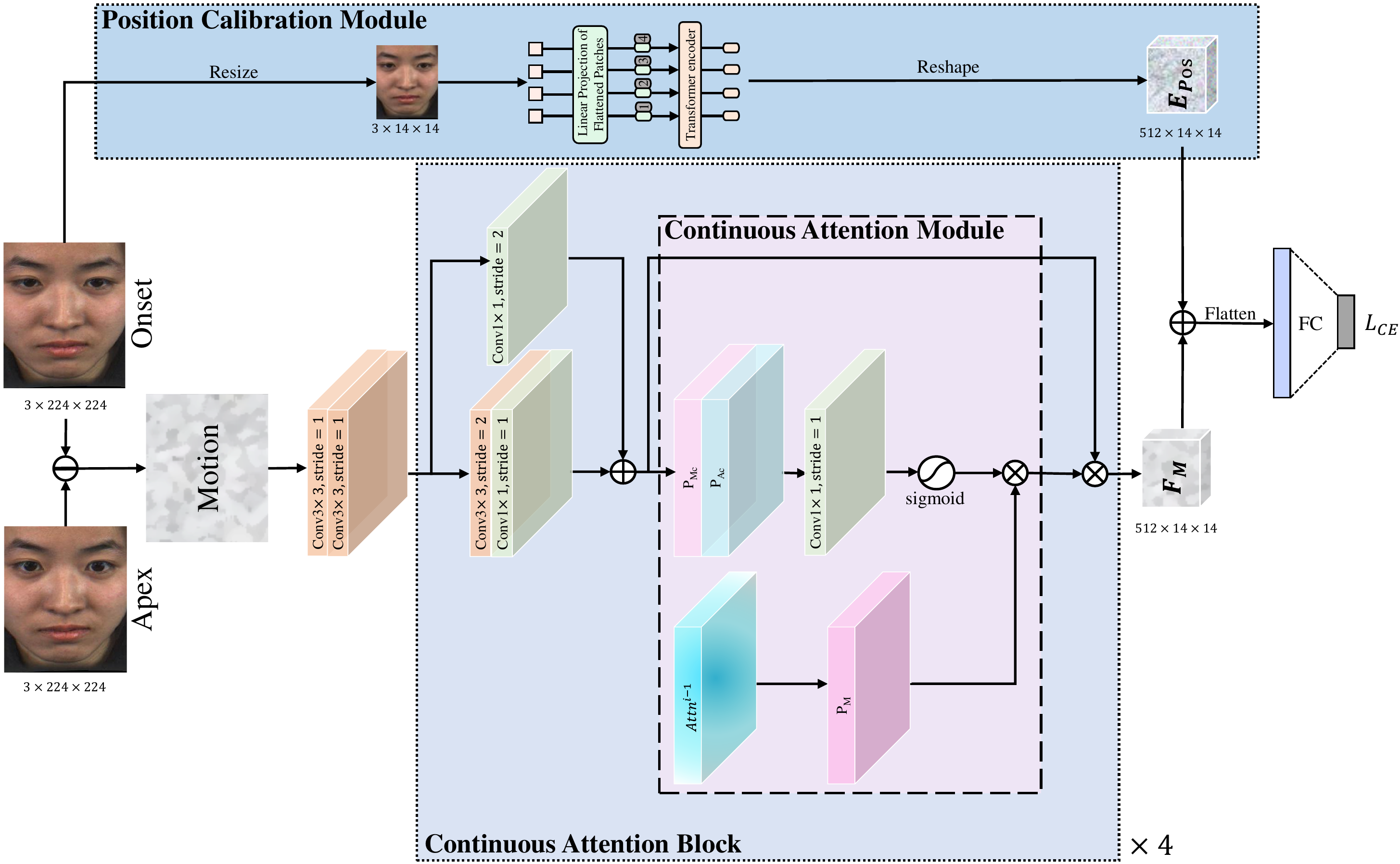}}
  \medskip
\caption{The pipeline of our MMNet. It contains two branches: the continuous attention block for extracting motion-pattern features and the position calibration module for locating the specific position of muscle motion. $L_{CE}$ stands for the cross-entropy loss function.}
\end{figure*}
\subsection{Problem Formulation and Overview}
Existing MER methods often focus on designing the structures of deep networks to improve the recognition accuracy, but usually neglect to find better ways to utilize the video frames of MEs. Most approaches input the original video frames with identity information into a single-branch network \cite{li2020LGCCon,lei2020Graph-TCN}, or extract handcrafted features (e.g., optical flow or LBP) from the frames and then fuse them by a multi-branch network \cite{liong2019STSTNet}. Since the input for the main branch contains the identity information of the sample to varying degrees, the network is likely to learn identity related features that have nothing to do with the MER tasks, especially when ME data are insufficient. To tackle this problem, we raise a novel two-branch MER paradigm. The main branch is designed to deal with the motion pattern, and the light-weight subbranch extracts facial position embeddings from the low-resolution onset frame that mainly contains the facial position information without any expression related information. In this way, even if the low-resolution onset frame contains certain identity information, since it is only fed into the light-weight subbranch, the whole network can still focus on learning the motion pattern rather than identity.

As seen in Figure 2, our proposed MMNet mainly consists of two branches. The main branch extracts the motion-pattern features with the difference between the apex frame and the onset frame as the input, and the subbranch is used to generate the facial position embeddings with low-resolution onset frame as the input. Finally, the facial position embeddings are added to the motion-pattern features for mapping the motion pattern to specific face areas.

\subsection{Continuous Attention Block}
Compared with macro expressions, MEs tend to have smaller muscle motions and more local active areas, which make it difficult for the traditional attention modules to accurately focus on the subtle facial muscle motions. To alleviate this, some existing works try to model the relationship between muscle motions and MEs \cite{xie2020AU-GACN} with the help of action unit (AU) labels \cite{friesen1978AUs}. However, it is still a huge challenge to obtain precise attention maps without introducing extra supervision.

To address the above issue, we devise a continuous attention block by introducing the attention maps of the previous layer as the prior knowledge to generate the attention maps of the current layer. Inspired by the spatial attention module of convolutional block attention module (CBAM) \cite{woo2018cbam} depicted in Figure 3(a), we utilize both max-pooling outputs and average-pooling outputs to calculate the spatial attention maps. As shown in Figure 3(b), we make the attention maps of the previous layer as the prior knowledge to obtain the attention maps of the current layer, and use a smaller convolution kernel (i.e., $1\times$$1$) to obtain more local attention maps. Formally, the CA module can be defined as,

\begin{equation}\label{1}
\resizebox{.91\linewidth}{!}{$
            \displaystyle
\begin{aligned}
Attn^{i}&={\rm M^{i}}(F_{conv}^{i}, Attn^{i-1})\\ &=\sigma (f_{1\times 1}^{i}([{\rm P_{Mc}}(F_{conv}^{i});{\rm P_{Ac}}(F_{conv}^{i})])) \bigotimes {\rm P_{M}}(Attn^{i-1}),
\end{aligned} $}
\end{equation}
\noindent with
\begin{equation}\label{1}
F_{conv}^{i} = f_{1\times 1}^{i}(f_{3\times 3}^{i}(F^{i}))+f_{1\times 1}^{i}(F^{i}),
\end{equation}
\noindent where $\rm M^{i}$ is the CA module of the $i^{th}$ CA block that expects to pay attention to the muscle motion areas and $Attn^{i-1}$ is the attention maps of the $(i-1)^{th}$ layer. $F_{conv}^{i}$$\in$$\rm{R}^{2C\times H\times W}$ denotes the features extracted by the first two convolutional layers of the $i^{th}$ layer as the input of CA module. $\sigma$ characterizes the sigmoid function. $f_{1\times 1}^{i}$ and $f_{3\times 3}^{i}$ represent a convolution operation with the kernel size of $1$ and $3$ from the $i^{th}$ layer, respectively. ${\rm P_{Mc}}(F_{conv}^{i})$$\in$$\rm{R}^{1\times H\times W}$ and ${\rm P_{Ac}}(F_{conv}^{i})$$\in$$\rm{R}^{1\times H\times W}$ separately describe the max-pooled features and average-pooled features across the channel. $\rm P_{M}$ indicates the max-pooling operation on the attention maps of $(i-1)^{th}$ layer to match the size of current layer attention maps, while $\bigotimes$ means the element-wise product for introducing the attention maps from last layer as prior knowledge. $F^{i}$$\in$$\rm{R}^{C\times 2H\times 2W}$ stands for the input of the $i^{th}$ CA block.

By associating the attention mechanism between adjacent layers, the CA block can gradually and robustly focus on the areas that have subtle motions, instead of focusing on different areas of the face in different layers, which may make networks learn both ME related areas and unrelated areas. We use four CA blocks to constitute the main branch of MMNet to learn the subtle muscle motion-pattern features $F_{M}$ with the size of $512$$\times$$14$$\times 14$. The CA block which consists of the CA module and two convolutional layers can be formulated as,
\begin{figure}[h]
	\centering
	\centerline{\includegraphics[width=8cm]{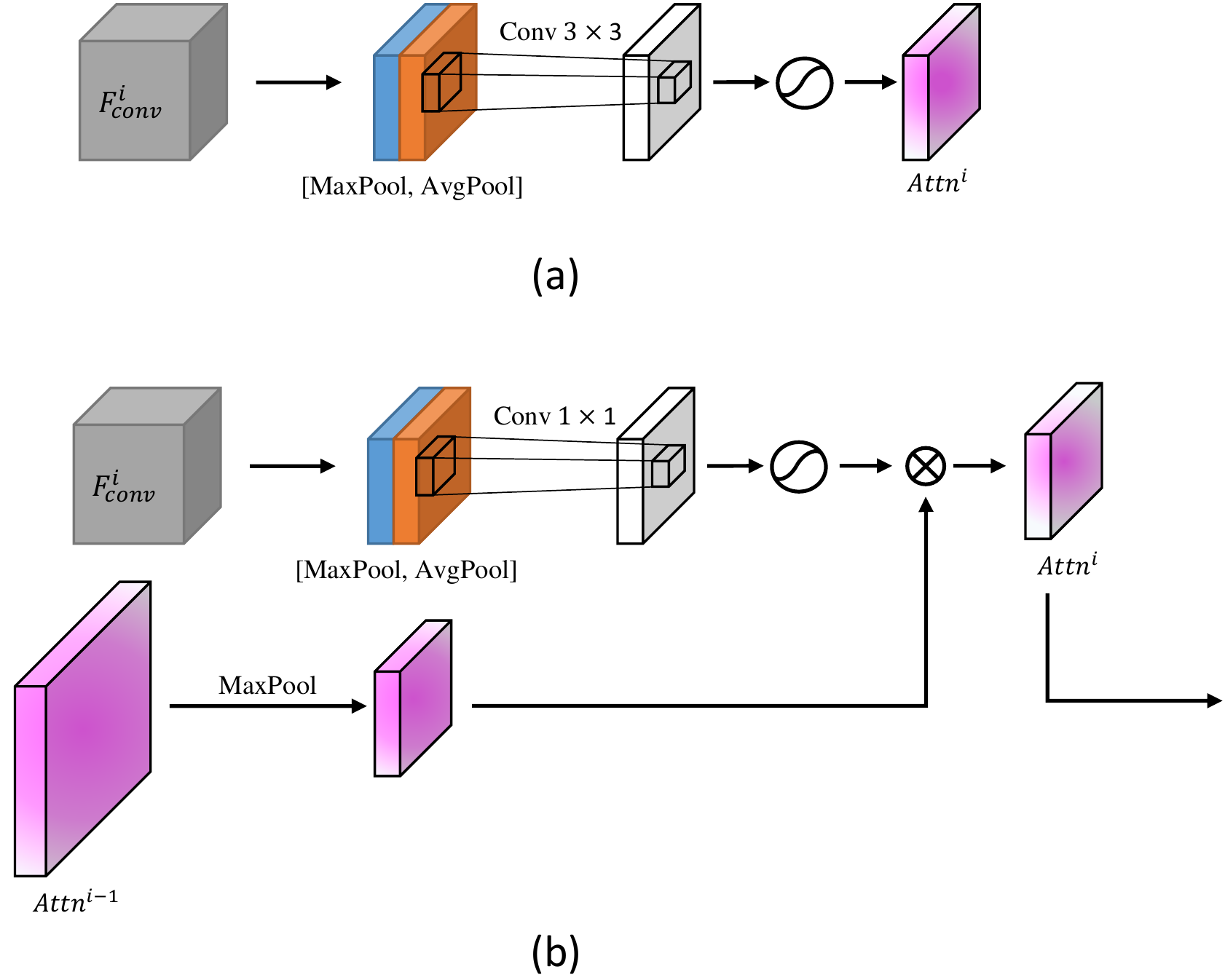}}	
	\caption{Diagram of the CBAM and CA modules. (a) Spatial attention module of CBAM module and (b) our CA module.}
	\label{figure1}
\end{figure}
\begin{equation}\label{1}
{\rm CA}(F^{i},Attn^{i-1})= F_{conv}^{i}\circledast {\rm M^{i}}(F_{conv}^{i},Attn^{i-1}),
\end{equation}

\noindent where $\rm CA$ represents the proposed continuous attention block and $\circledast$ stands for broadcast element-wise multiplication, which means each channel of $F_{conv}^{i}$$\in$$\rm{R}^{2C\times H\times W}$ will be multiplied by the spatial attention maps ${\rm M^{i}}(F_{conv}^{i},Attn^{i-1})$$\in$$\rm{R}^{1\times H\times W}$ to pay attention to the region of interest. The ablation experiments and visualization results in Figure 4 demonstrate that our CA block can help to concentrate on accurate ME related areas. As shown in Figure 2, we take the difference between the apex frame and the onset frame as the input of the main branch to learn the motion-pattern features.

\subsection{Position Calibration Module}
Due to the diverse appearances of different people in the micro-expression datasets, it is hard to strictly align all the faces because of various interpupillary distances, dissimilar nose sizes, etc. Therefore, the same face areas may correspond to different pixel positions of the image, which make it hard for the network to learn exactly where the subtle motion occurs. In order to accurately add position information to the motion-pattern features extracted by the main branch, we propose a position calibration module as the subbranch of MMNet to generate the facial position embeddings for mapping motion-pattern features to specific areas of the face.

Since the relative positions of the facial features are physically determined (e.g., the nose is usually located below the middle of the two eyes), modeling long-distance dependencies can effectively help locate the positions of various parts of the face and generate robust position embeddings. Recently, ViT applies self-attention mechanism to model long-distance dependencies and has achieved promising results on image classification tasks, while the convolutional neural networks (CNNs) often need many convolutional layers to obtain a global receptive field, which is not conducive to modeling facial position information. So, we introduce the PC module based on a shallow ViT. As shown in Figure 2, we utilize the difference between the apex frame and the onset frame to learn the motion-pattern features $F_{M}$ and the low-resolution onset frame to learn the facial position embeddings $E_{pos}$. Since we only need to learn the locations of salient areas (e.g., eyes, mouth, and nose), instead of the detailed texture related to the identity of samples (e.g., wrinkles and skin tone), we scale the onset frame to the same size of $14$$\times$$14$ as the input of subbranch matching the size of $F_{M}$. Then, we reshape the scaled onset frame into a sequence of 196 flattened 2D patches $I_{p}$ with the size of $1\times$$1\times$$3$, and map them through a trainable linear projection to get patch embeddings $E_{p}$ with $512$ dimensions, which matches the channel dimension of $F_{M}$. After adding the position embeddings of ViT to retain the position information as done in \cite{dosovitskiy2020vit}, these patch embeddings are sent to the transformer encoder to learn the relationship between patches. Finally, the output of ViT with the size of $196\times$$512$ is reshaped into $512\times$$14\times$$14$ to get $E_{pos}$ for position calibration. The position embeddings $E_{pos}$ are then added to the motion-pattern features $F_{M}$ for mapping the motion pattern to specific face areas for MER.

\section{Experiments}
\begin{table}
\centering

\begin{tabular}{|c|c|c|c|}
\hline
\diagbox{Label}{Dataset}& SAMM & CASME II& MMEW \\ 
\hline
 Happiness & 26 & 32&36 \\
\hline
 Anger & 57 & -- &--\\
 \hline
Contempt& 12  &--&--\\
\hline
 Disgust & -- & 63 &72\\
\hline
Repression& --  &27&--\\
\hline
Surprise& 15  &28&89\\
\hline
Others& 26  &99&66\\
\hline
\textbf{Total}& \textbf{136}  &\textbf{249}&\textbf{263}\\
\hline
\end{tabular}\caption{Summary of the data distributions for CASME II (five classes), SAMM (five classes), and MMEW (four classes).}
\label{tab:time}
\end{table}

\begin{table}
\centering

\begin{tabular}{|c|c|c|c|}
\hline
\diagbox{Label}{Dataset}& SAMM & CASME II \\ 
\hline
Positive & 26 & 32 \\
\hline
Negative & 92 & 96 \\
 \hline
Surprise& 15  &28\\
\hline

\textbf{Total}& \textbf{133}  &\textbf{156}\\
\hline
\end{tabular}\caption{Summary of the data distributions for CASME II (three classes) and SAMM (three classes).}
\label{tab:time}
\end{table}
To verify the effectiveness of our MMNet, we conduct extensive experiments on three popular micro-expression datasets including CASME II \cite{yan2014casmeii}, SAMM \cite{davison2016samm}, and MMEW \cite{ben2021mmew}. In this section, we first introduce these three datasets and the implementation details. Then, we explore the performance of the two branches of MMNet, respectively. Finally, we compare our method with several state-of-the-arts.
\begin{table*}[t]

  \centering

  \begin{threeparttable}

  \label{tab:performance_comparison}
    \begin{tabular*}{\hsize}{ccccccc}
    \toprule
    \multirow{2}{*}{Method}&
    \multicolumn{2}{c}{CASME II (5 classes) }&\multicolumn{2}{c}{ SAMM (5 classes)}&\multicolumn{2}{c}{MMEW (4 classes)}\cr
    \cmidrule(lr){2-3} \cmidrule(lr){4-5}\cmidrule{6-7}
    &Accuracy (\%)&F1-Score&Accuracy (\%)&F1-Score&Accuracy (\%)&F1-Score\cr
    \midrule
    Baseline (ResNet)&81.12&0.7582&73.53&0.6345&81.75&0.7921\cr
    CA block           &83.13&0.7891&76.47&0.6593&84.03&0.8206\cr
    PC module   &85.14&0.8362&77.21&0.6734&85.55&0.8402\cr
    \midrule
    PC module + CA block&\textbf{88.35}&\textbf{0.8676}&\textbf{80.14}&\textbf{0.7291}&\textbf{87.45}&\textbf{0.8635}\cr

    \bottomrule
    \end{tabular*}\caption{Evaluation of the continuous attention block and the position calibration module. The best results are highlighted in bold.}
    \end{threeparttable}
\end{table*}
\begin{table*}[htbp]
\label{2}

\begin{tabular*}{\hsize}{ccccccc}
    \toprule
    \multirow{2}{*}{Method}&
    \multicolumn{2}{c}{CASME II (5 classes) }&\multicolumn{2}{c}{SAMM (5 classes)}&\multicolumn{2}{c}{MMEW (4 classes)}\cr
    \cmidrule(lr){2-3} \cmidrule(lr){4-5}\cmidrule{6-7}
    &Accuracy (\%)&F1-Score&Accuracy (\%)&F1-Score&Accuracy (\%)&F1-Score\cr
    \midrule
    Independent attention&84.74&0.8216&77.21&0.7000&80.99&0.7941\cr
    Continuous attention          &\textbf{88.35}&\textbf{0.8676}&\textbf{80.14}&\textbf{0.7291}&\textbf{87.45}&\textbf{0.8635}\cr

    \bottomrule
\end{tabular*}\caption{Comparison of continuous attention module and independent attention module. The best results are highlighted in bold.}

\end{table*}
\subsection{Datasets}
\textbf{CASME II} \cite{yan2014casmeii} contains 256 micro-expression videos from 26 subjects with a cropped size of $280\times$$340$ at 200 fps. Consistent with most of previous methods, only samples of five prototypical expressions, i.e., happiness, disgust, repression, surprise, and others, are used. \textbf{SAMM} \cite{davison2016samm} has 159 micro-expression clips from 32 participants of 13 different ethnicities at 200 fps. Five expressions (happiness, anger, contempt, surprise, and others) are utilized for experiments. \textbf{MMEW} \cite{ben2021mmew} includes both macro- and micro-expressions sampled from the same subjects for researchers to explore the relationship between them. It contains 300 MEs and 900 macro-expression samples with a larger resolution ($1920\times$$1080$) at 90 fps. We use four expressions (happiness, surprise, disgust, and others) for ablation studies. Table 1 shows the data distributions for each class of CASME II (five classes), SAMM (five classes), and MMEW (four classes), while Table 2 gives the data distributions for each class of CASME II (three classes) and SAMM (three classes). First, we conduct ablation studies on these three datasets separately to verify the effectiveness of the CA block and PC module. Second, since the MMEW dataset was just released in 2021, there are currently few methods for comparison on this benchmark. So, we mainly compare MMNet with other state-of-the-art approaches on CASME II and SAMM datasets. Consistent with most of previous works, leave-one-subject-out (LOSO) cross-validation is employed in all the experiments, which means every subject is taken as a testing set in turn and the rest subjects as the training data. For all the experiments, the accuracy and F1-score are used for performance evaluation.

\subsection{Implementation Details}
In our experiments, we use $\it{Dlib}$ to detect the landmark points on the face and crop the images according to these landmark points. The cropped images on all the datasets are resized to 224 $\times$ 224. To avoid overfitting, we randomly pick a frame from four frames around the labeled onset and apex frames as the onset frame and apex frame for training. The horizontal flipping, random cropping, and color jittering are also employed. We use four CA blocks to constitute the main branch and a shallow ViT to build the subbranch. At the training stage, we adopt AdamW to optimize the MMNet with a batch size of 32. The learning rate is initialized to 0.0008, decreased at an exponential rate in 70 epochs for cross-entropy loss function. All the experiments are conducted on a single NVIDIA RTX 3070 card with PyTorch toolbox. 
\subsection{Ablation Studies}
\begin{table}[t]
\label{3}

\begin{tabular*}{\hsize}{c|cc|cc}

\toprule

 Setting&$N_{L}$&$N_{H}$&CASME II (\%)&SAMM (\%)\\
\midrule

i       &2     & 2&  86.35        & 79.41\\
ii      &2     & 4&\textbf{88.35} & 80.14\\
iii     &2     & 8&85.94          & 78.68\\
iv      &3     & 2&84.74          & \textbf{80.88}\\
v       &3     & 4&85.94          &80.14\\
vi      &3     & 8&84.34          &76.47\\

\bottomrule
\end{tabular*}\caption{ Ablation study w.r.t. number of heads and number of layers, performed on CASME II (5 classes) and SAMM (5 classes). $N_{L}$ represents the number of ViT encoder and $N_{H}$ stands for number of heads. Bold values correspond to the best performance.}

\end{table}

\paragraph{Effectiveness of the Two Branches in MMNet.}
To verify our CA block and PC module, we set a ResNet consisting of four building blocks which is used in ResNet18 as the baseline model for ablation studies. Then, we separately replace the building blocks of ResNet18 with the CA block and add the PC module to compare the proposed model with the baseline. As shown in Table 3, the CA block and PC module both can significantly improve the performance. Finally, when we use both the modules to build our MMNet, the results exceed the baseline by 7.23\%/10.94\%, 6.61\%/9.46\%, and 5.70\%/7.14\% on CASME II, SAMM, and MMEW datasets in terms of accuracy/F1-score.

\paragraph{Comparisons between Continuous Attention and Independent Attention.}
To illustrate the effectiveness of introducing attention maps of the previous layers as prior knowledge on generating precise attention maps for MER task, we compare the proposed continuous attention module with traditional independent attention module (i.e., spatial attention module of CBAM shown in Figure 3(a)), which generates the spatial attention maps independently in each layer. It is worthy to note that both settings add the PC module to generate the position embeddings. As shown in Table 4, the performance of CA module on the three datasets are significantly better than the independent attention module.

\paragraph{Impact of the number of layers and heads of PC module.}
Vision transformer encoder consists of $N_{L}$ identical layers. The multi-head self-attention in each layer enables the model to decompose the information into $N_{H}$ representation subspaces and jointly capture discriminative information at different positions. We explore the effects of different layer values $N_{L}$ and the number of heads $N_{H}$ on CASME II and SAMM datasets. Table 5 shows the performance comparison of different hyper-parameter settings of our method. We observe that smaller $N_{H}$ and $N_{L}$ tend to achieve better results. We think this is because larger models are more prone to overfitting. Therefore, we set the number of self-attention heads $N_{H}$ and the number of transformer encoder layers $N_{L}$ to 2 and 4 by default.
\begin{table}[t]
	\begin{center}
		\label{table2}
		\begin{tabular*}{\hsize}{cccc}
			\toprule  
			Method  & Cate & Acc (\%) &F1-Score  \\
			\midrule
			OFF-ApexNet \shortcite{gan2019off-apex}& 3&88.28&0.8697\\
            STSTNet  \shortcite{liong2019STSTNet}& 3&86.86&0.8382\\
            AU-GACN \shortcite{xie2020AU-GACN}& 3&71.20&0.3550\\
            MTMNet \shortcite{Xia2020MTMNet}& 3&75.60&0.7010\\
            MiNet\&MaNet \shortcite{xia2021MiNetMaNet}& 3&79.90&0.7590\\
            GACNN \shortcite{kumar2021GACNN}& 3&89.66&0.8695\\
            \midrule
            MMNet (Ours) & 3 & \textbf{95.51}&\textbf{0.9494} \\
            \midrule
            DSSN \shortcite{khor2019DSSN}& 5&71.19&0.7297\\
            TSCNN \shortcite{song2019TSCNN}& 5&80.97&0.8070\\
            Dynamic \shortcite{sun2020Dynamic}& 5&72.61&0.6700\\
            Graph-TCN \shortcite{lei2020Graph-TCN}& 5&73.98&0.7246\\
            SMA-STN \shortcite{liu2020SMA-STN}& 5&82.59&0.7946\\
            AU-GCN \shortcite{lei2021micro}& 5&74.27&0.7047\\
            GEME \shortcite{nie2021geme}& 5&75.20&0.7354\\
            MERSiamC3D \shortcite{zhao2021MERSiamC3D}& 5&81.89&0.8300\\
			\midrule

            MMNet (Ours) & 5 & \textbf{88.35}&\textbf{0.8676} \\
			\bottomrule
		\end{tabular*}\caption{Comparison results on CASME II dataset.``Cate" stands for the number of classes. The best accuracies are highlighted in bold.}

	\end{center}
\end{table}

\begin{table}[t]
	\begin{center}
		\label{table3}
		\begin{tabular*}{\hsize}{cccc}
			\toprule  
			Method  & Cate & Acc (\%) &F1-Score  \\
			\midrule
			OFF-ApexNet \shortcite{gan2019off-apex}& 3&68.18&0.5423\\
            STSTNet  \shortcite{liong2019STSTNet}& 3&68.10&0.6588\\
            AU-GACN \shortcite{xie2020AU-GACN}& 3&70.2&0.4330\\
            MTMNet \shortcite{Xia2020MTMNet}& 3&74.10&0.7360\\
            MiNet\&MaNet \shortcite{xia2021MiNetMaNet}& 3&76.70&0.7640\\
            GACNN \shortcite{kumar2021GACNN}& 3&88.72&0.8118\\
			\midrule
            MMNet (Ours) & 3 & $\textbf{90.22}$&\textbf{0.8391} \\
            \midrule
            DSSN \shortcite{khor2019DSSN}& 5&57.35&0.4644\\
            Graph-TCN \shortcite{lei2020Graph-TCN}& 5&75.00&0.6985\\
            SMA-STN \shortcite{liu2020SMA-STN}& 5&77.20&0.7033\\
            AU-GCN \shortcite{lei2021micro}& 5&74.26&0.7045\\
            GEME \shortcite{nie2021geme}& 5&55.38&0.4538\\
            MERSiamC3D \shortcite{zhao2021MERSiamC3D}& 5&68.75&0.6400\\
            \midrule

            MMNet (Ours) & 5 & $\textbf{80.14}$&\textbf{0.7291} \\
			\bottomrule
		\end{tabular*}\caption{Comparison results on SAMM dataset. ``Cate" indicates the number of classes. The best accuracies are highlighted in bold.}
	\end{center}
\end{table}
\begin{figure}[t!]
  \centering
  \centerline{\includegraphics[width=8.5cm]{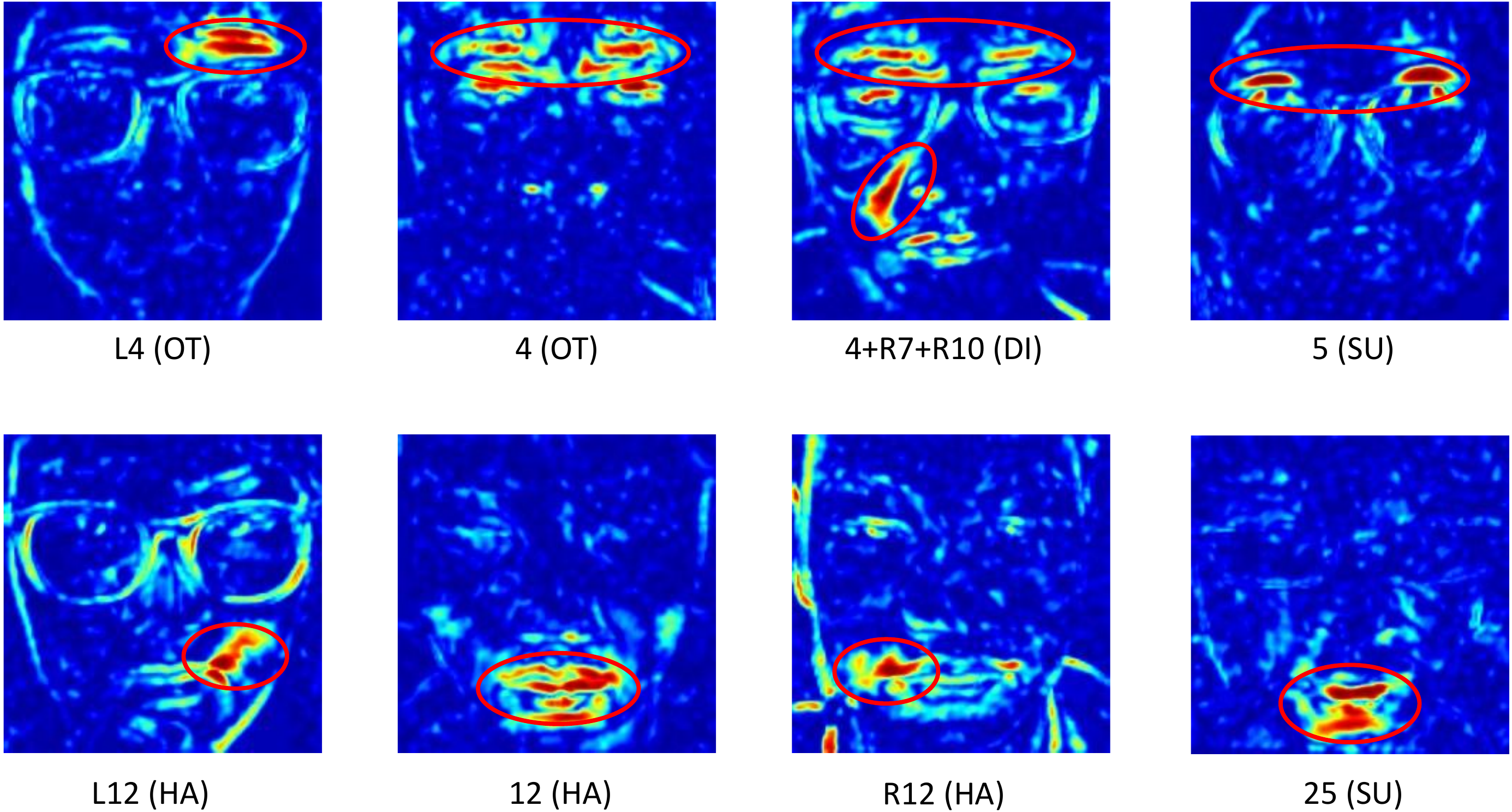}}

  \medskip

\caption{Visualization of the attention maps of several samples from MMEW dataset. The labels outside brackets represent the action unit label of the samples (e.g., L12 stands for the left corner of the lips, R12 represents the right corner of the lips, and 12 means both corners of the lips.), while the label inside the brackets is the expression type (e.g., SU, HA, DI, and OT represents surprise, happiness, disgust, and others, respectively).}
\end{figure}
\subsection{Comparison with State-of-the-arts}
We also compare our MMNet with several state-of-the-art methods. From Tables 6 and 7, we can see that our MMNet outperforms the best results of previous methods on every evaluation indicators. Specifically for three-class MER tasks, our method outperforms GACNN \cite{kumar2021GACNN} by 1.5\%/2.73\% and 5.85\%/7.99\% on SAMM and CASME II with respect to accuracy/F1-score. As for five classes, our MMNet exceeds SMA-STN \cite{liu2020SMA-STN} by 2.94\%/2.58\% and 5.76\%/7.3\% on SAMM and CASME II in terms of accuracy/F1-score. 

\subsection{Visualization}
In order to prove that our proposed continuous attention module can pay attention to the movements of tiny facial muscles, we visualize some attention maps of the first CA block in Figure 4. It can be easily seen that CA module can generate attention maps to help the network focus on where the facial muscle moves accurately. Specifically, the CA module concentrates on the upper end of the eyebrows for surprise samples and the corner of the mouth for happiness samples.

\section{Conclusion}
In this paper, we develop a new two-branch MER paradigm and make a realization of the new paradigm, called MMNet. Specifically, the main branch based on the proposed CA block focuses on learning motion-pattern features from the difference between the onset and apex frames, while the subbranch based on the PC module concentrates on generating facial position embeddings for position calibration. Extensive experiments illustrate that our MMNet outperforms state-of-the-arts by a large margin on CASME II and SAMM datasets.
\section*{Acknowledgements}
This work was supported by the JKW Research Funds under Grant 20-163-14-LZ-001-004-01. We acknowledge the support of GPU cluster built by MCC Lab of Information Science and Technology Institution, USTC.
\bibliographystyle{named}
\bibliography{ijcai22}

\end{document}